\documentclass{article}

\usepackage[final,nonatbib]{neurips_2024}
\usepackage[utf8]{inputenc} 
\usepackage[T1]{fontenc}    
\usepackage{hyperref}       
\usepackage{url}            
\usepackage{booktabs}       
\usepackage{amsfonts}       
\usepackage{nicefrac}       
\usepackage{microtype}      
\usepackage{wrapfig}

\usepackage{subfigure}
\usepackage[ruled]{algorithm2e}
\usepackage{bbm}
\usepackage{amsfonts}
\usepackage{epsfig}
\usepackage{graphicx}
\usepackage{amsmath,bm}
\usepackage{amssymb}
\usepackage{bbding}
\usepackage{color}
\usepackage{xcolor}
\usepackage{cleveref}
\usepackage{multirow}
\Crefname{figure}{Fig.}{Fig.}
\Crefname{equation}{Eq.}{Eq.}
\usepackage{etoolbox} 
\newcommand{\MyMapTemplatePrefixc}[4]{\expandafter#1\csname#3#4\endcsname{#2{#4}}} 
\forcsvlist{\MyMapTemplatePrefixc {\def} {\mathcal}{c}} {A,B,C,D,E,F,G,H,I,J,K,L,M,N,O,P,Q,R,S,T,U,V,W,X,Y,Z}  

\newcommand{\MyMapTemplatePrefixtb}[5]{\expandafter#1\csname#4#5\endcsname{#2{#3{#5}}}} 
\forcsvlist{\MyMapTemplatePrefixtb {\def} {\tilde}{\mathbf}{t}} {A,B,C,D,E,F,G,H,I,J,K,L,M,N,O,P,Q,R,S,T,U,V,W,X,Y,Z}  
\forcsvlist{\MyMapTemplatePrefixtb {\def} {\tilde}{\mathbf}{t}} {0,1,a,b,c,d,e,f,g,h,i,j,k,l,m,n,o,p,q,r,s,u,v,w,x,y,z}  

\newcommand{\MyMapTemplateNoPrefix}[3]{\expandafter#1\csname#3\endcsname{#2{#3}}}
\forcsvlist{\MyMapTemplateNoPrefix {\def} {\mathbf} } {0,1,a,b,c,d,e, f, g, h, i, j, k, l, m, n, o, p, q, r, u, v, w, x, y, z} 
\forcsvlist{\MyMapTemplateNoPrefix {\def} {\mathbf} } {A,B,C,D,E,F,G,H,I,J,K,L,M,N,O,P,Q,R,S,T,U,V,W,X,Y,Z}

\def\ie{\emph{i.e.}}

\usepackage{caption}
\title{FastDrag: Manipulate Anything in One Step}
\author{
Xuanjia Zhao$^1$,
Jian Guan$^{1,}$\thanks{Corresponding author: j.guan@hrbeu.edu.cn}
,
Congyi Fan$^1$,
Dongli Xu$^4$, \\
\textbf{Youtian Lin$^2$, Haiwei Pan$^1$, Pengming Feng$^3$}
\\ \\
$^1$College of Computer Science and Technology, Harbin Engineering University \\
$^2$School of Intelligence Science and Technology, Nanjing University \\
$^3$State Key Laboratory of Space-Ground Integrated Information Technology \\
$^4$Independent Researcher 
}

\begin{document}
\maketitle
\begin{abstract}
\label{abs}
Drag-based image editing using generative models provides precise control over image contents, enabling users to manipulate anything in an image with a few clicks.
However, prevailing methods typically adopt $n$-step iterations for latent semantic optimization to achieve drag-based image editing, which is time-consuming and limits practical applications.
In this paper, we introduce a novel one-step drag-based image editing method, \ie, FastDrag, to accelerate the editing process.
Central to our approach is a latent warpage function (LWF), which simulates the behavior of a stretched material to adjust the location of individual pixels within the latent space.
This innovation achieves one-step latent semantic optimization and hence significantly promotes editing speeds.
Meanwhile, null regions emerging after applying LWF are addressed by our proposed bilateral nearest neighbor interpolation~(BNNI) strategy. This strategy interpolates these regions using similar features from neighboring areas, thus enhancing semantic integrity.
Additionally, a consistency-preserving strategy is introduced to maintain the consistency between the edited and original images by adopting semantic information from the original image, saved as key and value pairs in self-attention module during diffusion inversion, to guide the diffusion sampling.
Our FastDrag is validated on the DragBench dataset, demonstrating substantial improvements in processing time over existing methods, while achieving enhanced editing performance. Project page: \url{https://fastdrag-site.github.io/}.
\end{abstract}
\section{Introduction}
\label{sec:1}
The drag editing paradigm~\cite{shi2023dragdiffusion,ling2023freedrag,mou2023dragondiffusion} leverages the unique properties of generative models to implement a point-interaction mode of image editing, referred to as drag-based image editing. Compared with text-based image editing methods~\cite{Mokady_2023_CVPR,ju2024direct,cho2024noise,xu2023infedit}, drag-based editing enables more precise spatial control over specific regions of the image while maintaining semantic logic coherence, drawing considerable attention from researchers.

\begin{figure}[t]
\vspace{-0.3cm}
\subfigure[Existing methods with $n$-step iterative optimization.]{
    \label{fig:compare_process_other}
    \includegraphics[width =0.47\textwidth]{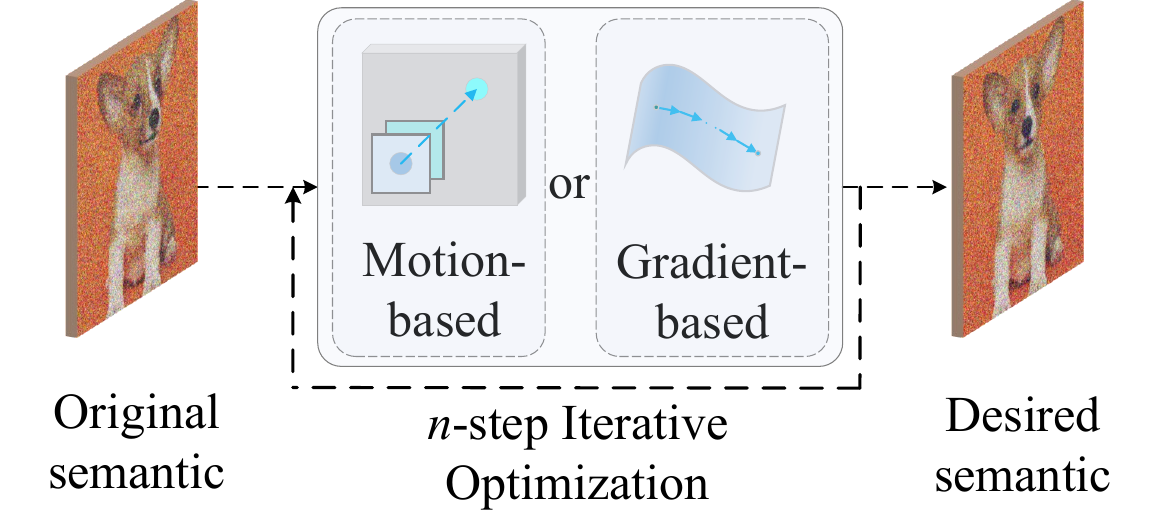}
    } 
\subfigure[FastDrag with one-step warpage optimization.]
{ 
    \includegraphics[width =0.47\textwidth]{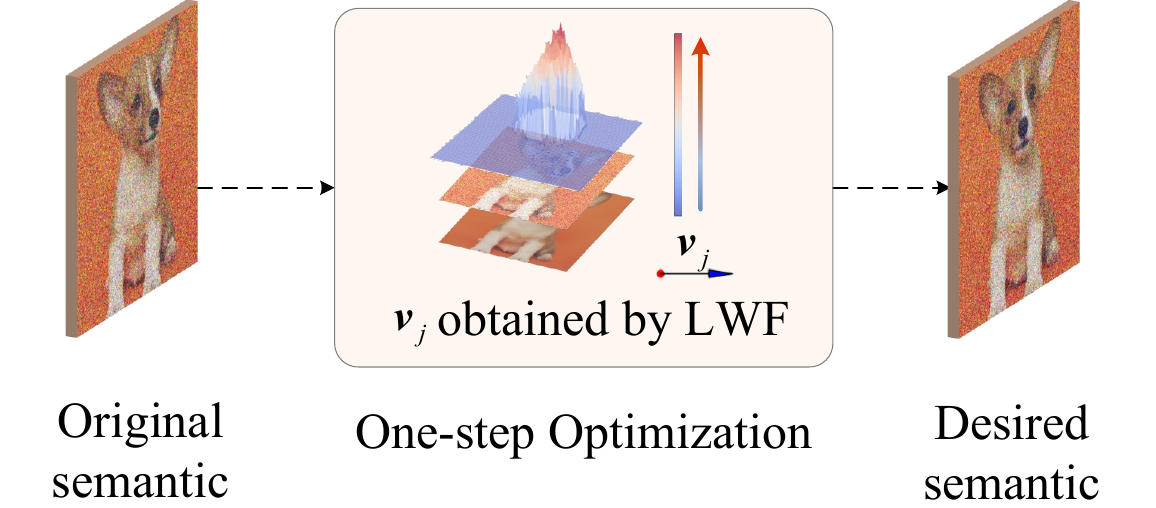}
\label{fig:compare_process_our}
}
\vspace{-0.25cm}
\caption{(a)~Existing methods usually require multiple iterations to transform an image from its original semantic to desired semantic; (b)~Our method utilizes latent warpage function (LWF) to calculate the warpage vectors (\ie, $\boldsymbol{v}_j$) to move each individual pixel on feature map and achieve semantic optimization in one step.}
\vspace{-0.5cm}
\end{figure}

However, existing methods typically involve $n$-step iterative semantic optimization in latent space to obtain optimized latent with desired semantic based on the user-provided drag instructions. 
They focus on optimizing a small region of the image at each step, requiring $n$ small-scale and short-distance adjustments to achieve overall latent optimization, leading to a significant amount of time.
These optimization approaches can be categorized primarily into motion-based~\cite{pan2023drag, shi2023dragdiffusion, ling2023freedrag, liu2024drag, zhang2024gooddrag, Cui2024StableDragSD} and gradient-based~\cite{mou2023dragondiffusion, mou2024diffeditor} $n$-step iterative optimizations, as shown in \Cref{fig:compare_process_other}. $n$-step iterations in motion-based methods are necessary to avoid abrupt changes in the latent space, preventing image distortions and ensuring a stable optimization process. This is exemplified in DragDiffusion~\cite{shi2023dragdiffusion} and GoodDrag~\cite{zhang2024gooddrag}, which require 70 to 80 iterations of point tracking and motion supervision for optimization. In addition, gradient-based methods align sampling results with the drag instructions through gradient guidance~\cite{epstein2023diffusion}. In this way, they also require multiple steps due to the optimizer~\cite{kingma2014adam, Robbins1951ASA} needing multiple iterations for non-convex optimization.  For instance, DragonDiffusion~\cite{mou2023dragondiffusion} requires around 50 gradient steps to accomplish the latent optimization. Therefore, existing drag-based image editing methods often suffer from significant time consumption due to $n$-step iterations required for latent semantic optimization, thus limiting the practical applications. 

To this end, we present a novel one-step drag-based image editing method based on diffusion, \ie, FastDrag, which significantly accelerates editing speeds while maintaining the quality and precision of drag operations. Specifically, a novel one-step warpage optimization strategy is proposed to accelerate editing speeds, which can achieve the latent semantic optimization in a single step with an elaborately designed latent warpage function (LWF), instead of using motion or gradient-based $n$-step optimizations, as illustrated in \Cref{fig:compare_process_our}.
By simulating strain patterns in stretched materials, we treat drag instructions on the noisy latent as external forces stretching a material, and introduce a stretch factor in LWF, which enables the LWF to generate warpage vectors to adjust the position of individual pixels on the noisy latent with a simple latent relocation operation, thus achieving one-step optimization for drag-based editing. Meanwhile, a bilateral nearest neighbor interpolation (BNNI) strategy is proposed to enhance the semantic integrity of the edited content, by interpolating null values using similar features from their neighboring areas to address semantic losses caused by null regions emerging after latent relocation operation, thus enhancing the quality of the drag editing. 

Additionally, a consistency-preserving strategy is introduced to maintain the consistency of the edited image, which adopts the original image information saved in diffusion inversion (\ie, key and value pairs of self-attention in the U-Net structure of diffusion model) to guide the diffusion sampling for desired image reconstruction, thus achieving precise editing effect.
To further reduce time consumption for inversion and sampling, the latent consistency model (LCM)~\cite{luo2023lcm} is employed in the U-Net architecture of our diffusion-based FastDrag. Therefore, our FastDrag can significantly accelerate editing speeds while ensuring the quality of drag effects.

Experiments on DragBench demonstrate that the proposed FastDrag is the fastest drag-based editing method, which is nearly 700\% faster than the fastest existing method (\ie, DiffEditor~\cite{mou2024diffeditor}), and 2800\% faster than the typical baseline method (\ie, DragDiffusion~\cite{shi2023dragdiffusion}), with comparable editing performance.  We also conduct rigorous ablation studies to validate the strategies used in FastDrag.

\textbf{Contributions:}
1) We propose a novel drag-based image editing approach based on diffusion\, \ie, FastDrag, where a LWF strategy is proposed to achieve one-step semantic optimization, tremendously enhancing the editing efficiency.
2) We propose a novel interpolation method (\ie, BNNI), which effectively addresses the issue of null regions, thereby enhancing the semantic integrity of the edited content.
3) We introduce a consistency-preserving strategy to maintain the image consistency during editing process.
\section{Related Work}
\label{sec:2}
\subsection{Text-based Image Editing}
Text-based image editing has seen significant advancements, allowing users to manipulate images through natural language instructions. DiffusionCLIP~\cite{9879284} adopts contrastive language-image pretraining (CLIP)~\cite{pmlrv139radford21a} for diffusion process fine-tuning to enhance the diffusion model, enabling high-quality zero-shot image editing. The study in~\cite{hertz2022prompttoprompt} manipulates the cross-attention maps within the diffusion process and achieves text-based image editing. Imagic~\cite{Kawar_2023_CVPR} further enhances these methods by optimizing text embeddings and using text-driven fine-tuning of the diffusion model, enabling complex semantic editing of images. InstructPix2Pix~\cite{Brooks_2023_CVPR} leverages a pre-trained large language model combined with a text-to-image model to generate training data for a conditional diffusion model, allowing it to edit images directly based on textual instructions during forward propagation. Moreover, Null-text Inversion~\cite{Mokady_2023_CVPR} enhances text-based image editing by optimizing the default null-text embeddings to achieve desired image editing. Although text-based image editing methods enable the manipulation of image content using natural language description, they often lack the precision and explicit control provided by drag-based image editing.
\subsection{Drag-based Image Editing}
\label{sec:2.2}
Drag-based image editing achieves precise spatial control over specific regions of the image based on user-provided drag instructions. Existing drag-based image editing methods generally rely on $n$-step latent semantic optimization in latent space to achieve image editing. These methods fall into two main categories: motion-based~\cite{pan2023drag, shi2023dragdiffusion, zhang2024gooddrag, Cui2024StableDragSD, liu2024drag, ling2023freedrag, hou2024easydrag} and gradient-based~\cite{mou2023dragondiffusion, mou2024diffeditor} optimizations. For example, DragGAN~\cite{pan2023drag} employs generative adversarial network (GAN) for drag-based image editing with iterative point tracking and motion supervision steps. However, the image quality of the methods using GAN for image generation is worse than diffusion models~\cite{Dhariwal2021DiffusionMB}. Therefore, a series of diffusion-based methods have been proposed for drag-based image editing. For instance, DragDiffusion~\cite{shi2023dragdiffusion} employs iterative point tracking and motion supervision for latent semantic optimization to achieve drag-based editing. Building on this foundation, GoodDrag~\cite{zhang2024gooddrag}, StableDrag~\cite{Cui2024StableDragSD}, DragNoise~\cite{liu2024drag}, and FreeDrag~\cite{ling2023freedrag} have made significant improvements to the motion-based methods. Without coincidence, by utilizing feature correspondences, DragonDiffusion~\cite{mou2023dragondiffusion} and its improved version DiffEditor~\cite{mou2024diffeditor} formulate an energy function that conforms to the desired editing results, thereby transforming the image editing task into a gradient-based process that enables drag-based editing. However, these methods inherently require $n$-step iterations for latent optimization, which significantly increases the time consumption. Although SDEDrag~\cite{nie2024the} does not require $n$-step iterative optimization, it is still time-consuming due to the stochastic differential equation (SDE) process for diffusion. In addition, while EasyDrag~\cite{hou2024easydrag} offers user-friendship editing, its requirement for over 24GB of memory (\ie, a 3090 GPU) limits its broad applicability. To this end, based on latent diffusion model (LDM)~\cite{Rombach2021HighResolutionIS}, we propose a novel one-step optimization method that substantially accelerates the image editing speeds.
\section{Proposed Method}
\label{sec:3}
\begin{figure}[t]
    \centering
    \vspace{-0.7cm}
\includegraphics[width=.98\textwidth]{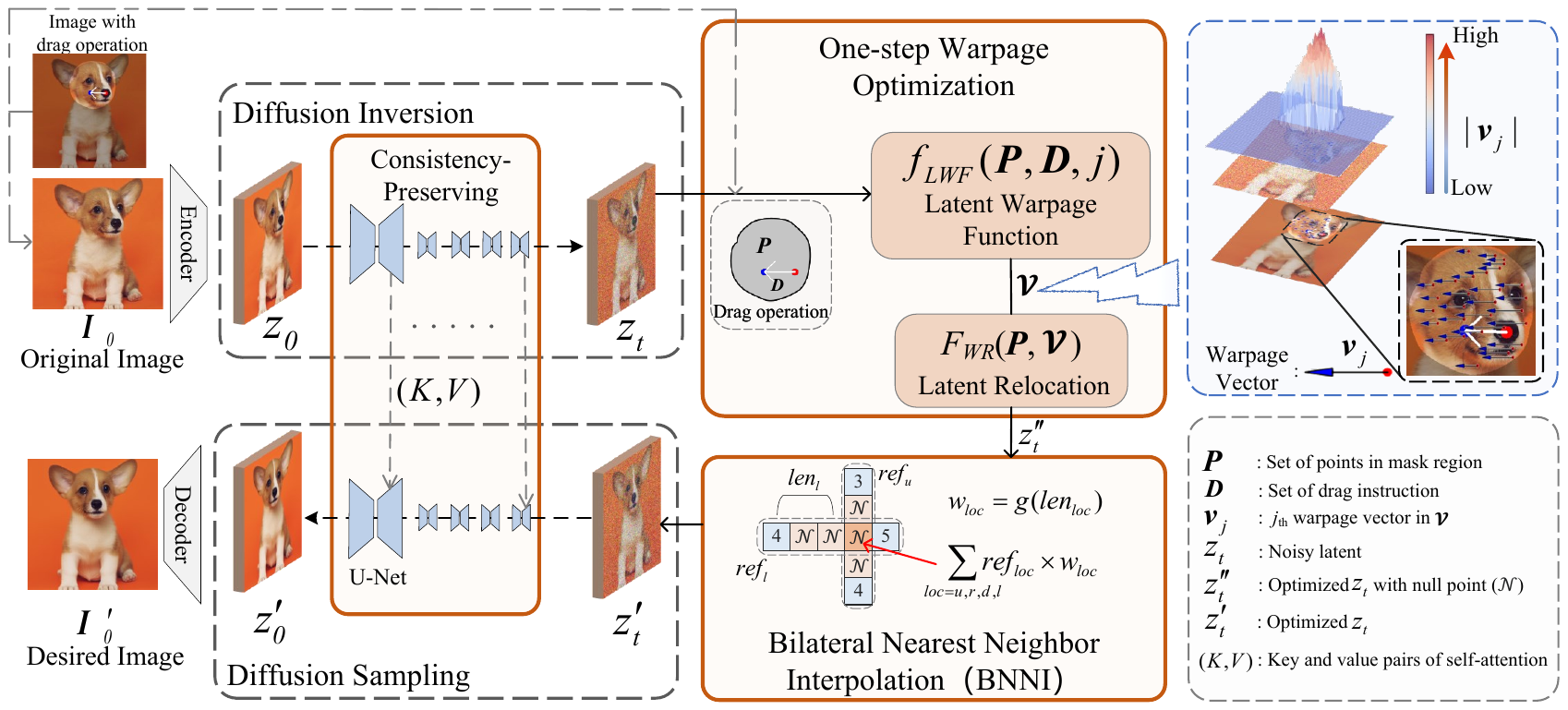}
    \caption{Overall framework of FastDrag with four phases: diffusion inversion, diffusion sampling, one-step warpage optimization and BNNI. Diffusion inversion yields a noisy latent $\boldsymbol{z}_t$ and diffusion sampling reconstructs the image from the optimized noisy latent $\boldsymbol{z}'_t$. One-step warpage optimization is used for noisy latent optimization, where LWF is proposed to generate warpage vectors to adjust the location of individual pixels on the noisy latent with a simple latent relocation operation. BNNI is used to enhance the semantic integrity of noisy latent. A consistency-preserving strategy is introduced to maintain the consistency between original image and edited image.}
    \label{fig1: overall-method-structure}
    \vspace{-0.6cm}
\end{figure}

FastDrag is based on LDM~\cite{Rombach2021HighResolutionIS} to achieve drag-based image editing across four phases. The overall framework is given in~\Cref{fig1: overall-method-structure}, and the detailed description of strategies in FastDrag are presented as follows:~(1)~Initially, FastDrag is based on a traditional image editing framework including diffusion inversion and sampling processes, which will be elaborated in Sec.~\ref{sec:3.1}. 
(2)~The core phase in Sec.~\ref{sec:3.2} is a one-step warpage optimization, employing LWF and a latent relocation operation to simulate the behavior of stretched material, allowing for fast semantic optimization.
(3)~BNNI is then applied in Sec.~\ref{sec:3.3} to enhance the semantic integrity of the edited content, by interpolating the null regions emerging after the one-step warpage optimization.
(4)~The consistency-preserving strategy is introduced in Sec.~\ref{sec:3.4} to maintain the desired image consistency with original image, by utilizing the key and value of self-attention in inversion to guide the sampling.
\subsection{Diffusion-based Image Editing}
\label{sec:3.1}
Similar to most existing drag editing methods~\cite{shi2023dragdiffusion, zhang2024gooddrag, mou2023dragondiffusion}, FastDrag is also built upon diffusion model (\ie, LDM), including diffusion inversion and diffusion sampling.

\noindent \textbf{Diffusion Inversion}~\cite{song2021denoising} is about mapping a given image to its corresponding noisy latent representation in the model's latent space. We perform semantic optimization on the noisy latent $\boldsymbol{z}_t \in \mathbb{R}^{w\times{h}\times{c}}$, due to it still captures the main semantic features of the image but is perturbed by noise, making it suitable as a starting point for controlled modifications and sampling~\cite{shi2023dragdiffusion}. Here, $w$, $h$, $c$ represent the width, height and channel of $\boldsymbol{z}_t$, respectively. This process for a latent variable at diffusion step $t$ can be expressed as:
\begin{align}
\label{eq:1}
\boldsymbol{z}_t=\frac{\sqrt{{\alpha}_t}}{\sqrt{\alpha_{t-1}}}(\boldsymbol{z}_{t-1}-\sqrt{1-{\alpha_{t-1}}} \cdot \boldsymbol{\epsilon}_t)+\sqrt{1-{\alpha_t}} \cdot \boldsymbol{\epsilon}_t,
\end{align}
where $z_0=\mathcal{E}(\boldsymbol{I}_0)$ denotes the initial latent of the original image $\boldsymbol{I}_0$ from the encoder~\cite{Kingma2013AutoEncodingVB} $\mathcal{E}(\cdot)$.
$\alpha_t$ is the noise variance at diffusion step $t$, and ${\boldsymbol{\epsilon}_t}$ is the noise predicted by U-Net. 
Subsequently, we perform a one-step warpage optimization on $\boldsymbol{z}_t$ in Sec.~\ref{sec:3.2}. 

\noindent \textbf{Diffusion Sampling} reconstructs the image from the optimized noisy latent $\boldsymbol{z}_t'$ by progressively denoising it to the desired latent $\boldsymbol{z}_0'$. This sampling process can be formulated as:
\begin{align}
\label{eq:2}
\boldsymbol{z}_{t-1}'=
\sqrt{{\alpha_{t-1}}} \cdot 
\left(\frac{\boldsymbol{z}_t'-\sqrt{1-{\alpha_t}} \cdot \boldsymbol{\epsilon}_t}{\sqrt{{\alpha_t}}}\right)+\sqrt{1-{\alpha_{t-1}} - \sigma^2} \cdot \boldsymbol{\epsilon}_t+\sigma^2 \cdot \boldsymbol{\epsilon},
\end{align}
where $\boldsymbol{\epsilon}$ is the Gaussian noise and $\sigma$ denotes the noise level. By iterating the process from $t$ to 1, $\boldsymbol{z}_0'$ is reconstructed, and the desired image can be obtained by $\boldsymbol{I}_0'=\mathcal{D}(\boldsymbol{z}_0')$, with $\mathcal{D}(\cdot)$ being the decoder~\cite{Kingma2013AutoEncodingVB}.
\subsection{One-step Warpage Optimization} 
\label{sec:3.2}
Building upon the phases in Sec.\ref{sec:3.1}, we propose a one-step warpage optimization for fast drag-based image editing.
The core idea involves simulating strain patterns in stretched materials, where drag instructions on the noisy latent are interpreted as external forces stretching the material. This enables us to adjust the position of individual pixels on the noisy latent, optimizing the semantic of noisy latent in one step, thus achieving extremely fast drag-based editing speeds.
To this end, we design the LWF in Sec.~\ref{sec:3.2.1} to obtain warpage vector, which is utilized by a straightforward latent relocation operation in Sec.~\ref{sec:3.2.2} to adjust the position of individual pixels on the noisy latent.
\subsubsection{Warpage Vector Calculation using LWF}
\label{sec:3.2.1}
\begin{wrapfigure}{r}{0.45\textwidth}
\setlength{\abovecaptionskip}{0.cm}
\setlength{\belowcaptionskip}{0.cm}
    \vspace{-0.6cm}
    \begin{center}
\includegraphics[width=.45\textwidth]{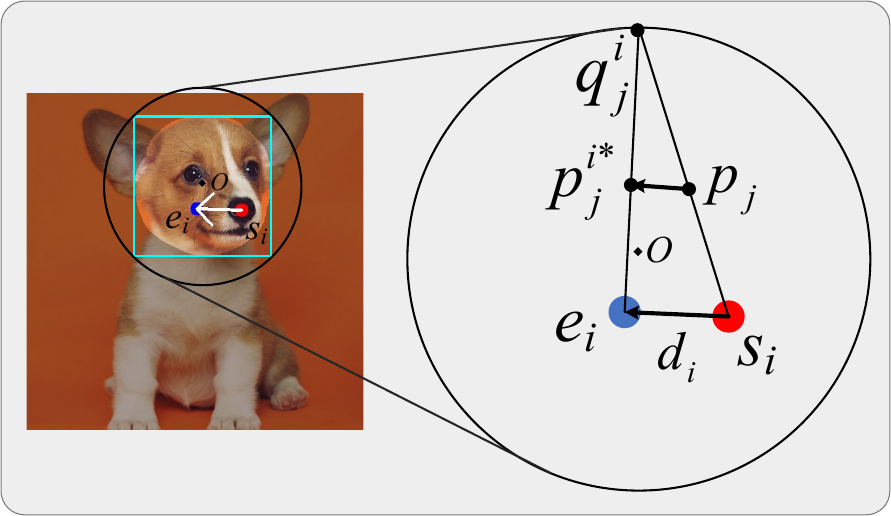}
    \end{center}
    \caption{Geometric representation of $\boldsymbol{v}_j^{i*}$.
    Circle $O$ is the circumscribed circle of the circumscribed rectangle enclosing the mask's shape.
    $p_j$ is the feature point requiring relocation, and $p_j^{i*}$ is its new position following the drag instruction $\boldsymbol{d}_{i}$}
    \label{stretch-facter}
    \vspace{-01cm}
\end{wrapfigure}

In drag-based image editing, each drag instruction $\boldsymbol{d}_{i}$ in a set of $k$ drag instructions $\boldsymbol{D} = \{\boldsymbol{d}_{i} \mid i = 1, \ldots, k;k\in\mathbb{Z}\}$ can simultaneously influence a feature point $p_j$ on the mask region $\boldsymbol{P} = \{p_j \mid j = 1, \ldots, m; m \in \mathbb{Z}\}$ provided by the user. 
As shown in~\Cref{stretch-facter}, the mask region is represented by the brighter areas in the image, indicating the specific image area to be edited.
To get a uniquely determined vector, \ie, warpage vector $\boldsymbol{{v}_j}$ to adjust the position of feature point $p_j$ (will be discussed in Sec.\ref{sec:3.2.2}), we propose a latent warpage function $f_{LWF}(\cdot)$ to aggregate multiple component warpage vectors caused by different drag instructions, \ie, $\boldsymbol{v}_j^{i*}$, with balanced weights to avoid deviating from the desired drag effect. The function is given as follows:
\begin{equation}
\label{eq:lwf}
\begin{aligned}
\boldsymbol{v}_j = f_{LWF}( \boldsymbol{P},\boldsymbol{D},j) =\sum_i^k{w_j^i \cdot \boldsymbol{v}_j^{i*} }, 
\end{aligned}
\end{equation}
where $w_j^i$ is the normalization weight for component warpage vector $\boldsymbol{v}_j^{i*}$.
Here, drag instruction $\boldsymbol{d}_{i}$ is considered as a vector form handle point $s_i$ to target point $e_i$.
During dragging, we aim for the semantic changes around the handle point $s_i$ to be determined by the corresponding drag instruction $\boldsymbol{d}_{i}$, rather than other drag instructions far from the $s_i$. Therefore,  $w_j^i$ is calculated as follows:
\begin{align}
w_j^i &= \frac{1 / {|p_js_i|}} { \sum_i^k{(1 / {|p_js_i|)}} },
\end{align}
where $s_i$ is considered as the ``point of force'' of $\boldsymbol{d}_{i}$, and the weight $w_j^i$ is inversely proportional to the Euclidean distance from $s_i$ to $p_j$.

It is worth noting that under an external force, the magnitude of component forces at each position within the material is inversely proportional to the distance from the force point, while the movement direction at each position typically aligns with the direction of the applied force~\cite{Naylor_1969}. Similarly, the component warpage vector $\boldsymbol{{v}}_j^{i*}$ on each $p_j$ aligns with the direction of drag instruction $\boldsymbol{d}_{i}$, and magnitudes of $\boldsymbol{v}_j^{i*}$ are inversely proportional to the distance from $s_i$. 
Hence, $\boldsymbol{{v}}_j^{i*}$ can be simplified as:
\begin{align}
\boldsymbol{{v}}_j^{i*} =  \lambda_j^i \cdot \boldsymbol{d}_i,
\label{effect of factor}
\end{align}
where $\lambda_j^i$ is the stretch factor that denotes the  proportion between $\boldsymbol{{v}}_j^{i*}$ and $\boldsymbol{d}_{i}$.

To appropriately obtain the stretch factor $\lambda_j^i$ and facilitate the calculation, we delve into the geometric representation of the component warpage vector $\boldsymbol{v}_j^{i*}$. 
As shown in Fig.~\ref{stretch-facter}, $\boldsymbol{v}_j^{i*}$ can be depicted as the guidance vector from point $p_j$ to point $p_j^{i\ast}$, where $p_j^{i\ast}$ is the expected new position of $p_j$ under the drag effect of $\boldsymbol{d}_{i}$.
Recognizing that the content near to mask edge should remain unaltered, we define a reference circle $O$ where every $\boldsymbol{v}_j^{i*}$ will gradually reduce to $0$ as $p_j$ approaches the circle.
Consequently, since $\boldsymbol{v}_j^{i*}$ and $\boldsymbol{d}_{i}$ are parallel, magnitudes of $\boldsymbol{v}_j^{i*}$ are inversely proportional to the distance from $s_i$ and $\boldsymbol{v}_j^{i*}$ is reduced to $0$ on circle $O$, the extended lines from $s_i p_j$ and $e_ip_j^{i\ast}$ will intersect at $q_j^i$  on circle $O$.
Hence, based on the~\Cref{effect of factor} and the geometric principle in~\Cref{stretch-facter}, we calculate $\lambda_j^i$ as follows:
\begin{align}
\label{eq:4}
\lambda_j^i = \frac{|\boldsymbol{{v}}_j^{i*}|}{|\boldsymbol{d}_i|} = \frac{|\overrightarrow{p_j p_j^{i\ast}}|}{|\overrightarrow{s_i e_i}|} = \frac{|p_j q_j^i|}{|s_i q_j^i|}.
\end{align}
Finally, we obtain the warpage vector $\boldsymbol{v}_j$ using only $\boldsymbol{d}_i$ and two factors as follows:
\begin{equation}
f_{LWF}( \boldsymbol{P},\boldsymbol{D},j) =\sum_i^k{w_j^i \cdot \lambda_j^i \cdot \boldsymbol{d}_i}
\end{equation}
Note that, for the special application of drag-based editing, such as object moving  as shown in Fig.~\ref{ablution_inversion}, drag editing is degenerated to a mask region shifting operation, requiring the spatial semantics of the mask region to remain unchanged. In that case, we only process a single drag instruction, and all component drag effects will be set equal to the warpage vector, \ie, $\boldsymbol{v}_j=\boldsymbol{d}_{1}$ and  $\boldsymbol{D}=\{\boldsymbol{d}_{1}$\}.
\subsubsection{Latent Relocation with Warpage Vector}
\label{sec:3.2.2}
Consequently, we utilize the warpage vector $v_j$ to adjust the position of feature point $p_j$ via a latent relocation operation $F_{WR}$, achieving the semantic optimization of noisy latent for drag-based editing. Establishing a Cartesian coordinate system on the latent space, let $(x_{p_j}, y_{p_j})$ denote the position of point $p_j \in \boldsymbol{P}$ within this coordinate system.
The new location of the point $p_j^*$ after applying the vector $\boldsymbol{v}_j = (v_{j}^{x}, v_{j}^{y}) $ can be written as:
\begin{equation}
    (x_{p_j}^{*}, y_{p_j}^{*}) = (x_{p_j}, y_{p_j}) + (v_{j}^{x}, v_{j}^{y})
\end{equation}
Then the new coordinates set $\boldsymbol{C}$ of all feature points in $\boldsymbol{P}$ can be written as:
\begin{equation}
\label{eq:FWR}
\begin{aligned}
\boldsymbol{C} = F_{WR}(\boldsymbol{P}, \boldsymbol{\mathcal{V}})
= \{(x_{p_j}^{*}, y_{p_j}^{*}) | (x_{p_j}^{*}, y_{p_j}^{*})  = (x_{p_j}, y_{p_j}) + (v_{j}^{x}, v_{j}^{y}); j=1,\cdots,m \},
\end{aligned}
\end{equation}
where $\boldsymbol{\mathcal{V}} =\{\boldsymbol{v}_j|j=1,\cdots, m,m\in\mathbb{Z}\}$.
If $(x_{p_j}, y_{p_j})$ has already been a new position for a feature point, it no longer serves as a new position for any other points. Consequently, by assigning corresponding values to these new positions, the optimized noisy latent $\boldsymbol{z}''_t$ can be obtained as shown in the following equation:
\begin{equation}
{\boldsymbol{z}''_t}_{(x_{p_j}+v_{j}^{x}, y_{p_j}+v_{j}^{y})} = {\boldsymbol{z}_{t}}_{(x_{p_j}, y_{p_j})}
\end{equation}
In essence, the latent relocation operation optimizes semantics efficiently by utilizing the LWF-generated warpage vector, eliminating the need for iterative optimization.

However, as certain positions in the noisy latent may not be occupied by other feature points,  $\boldsymbol{z}''_t$ obtained from one-step warpage optimization may contain regions with null values as shown in Fig.~\ref{interpolation}, leading to semantic losses that can adversely impact the drag result.
We address this issue in Sec.~\ref{sec:3.3}.
\subsection{Bilateral Nearest Neighbor Interpolation}
\label{sec:3.3}
\begin{figure}[t]
    \hspace{-0.5cm}
    \begin{minipage}{.62\textwidth}
        \vspace{-10pt}
        \centering
        \includegraphics[width=.98\textwidth]{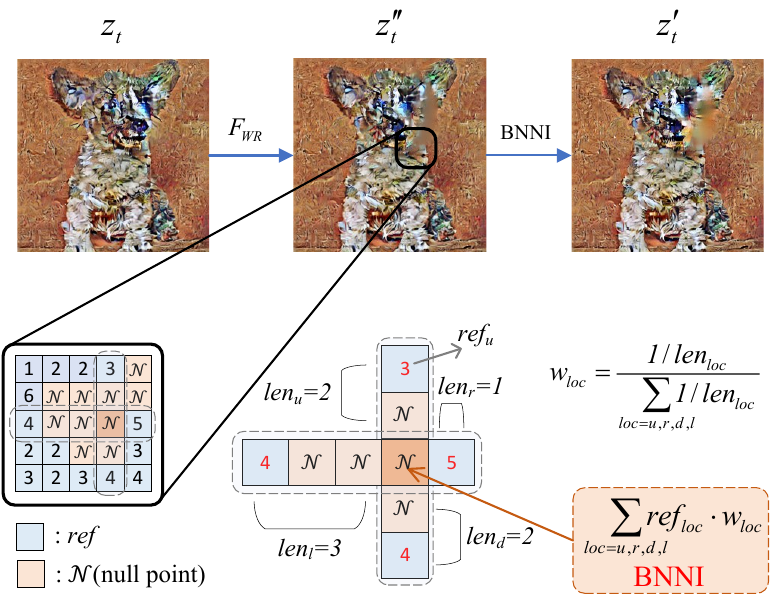}
        \caption{Illustration of bilateral nearest neighbor interpolation.}
        \label{interpolation}
    \end{minipage}
    \hspace{+10pt}
    \begin{minipage}{.38\textwidth}
        \centering
        \includegraphics[width=.98\textwidth]{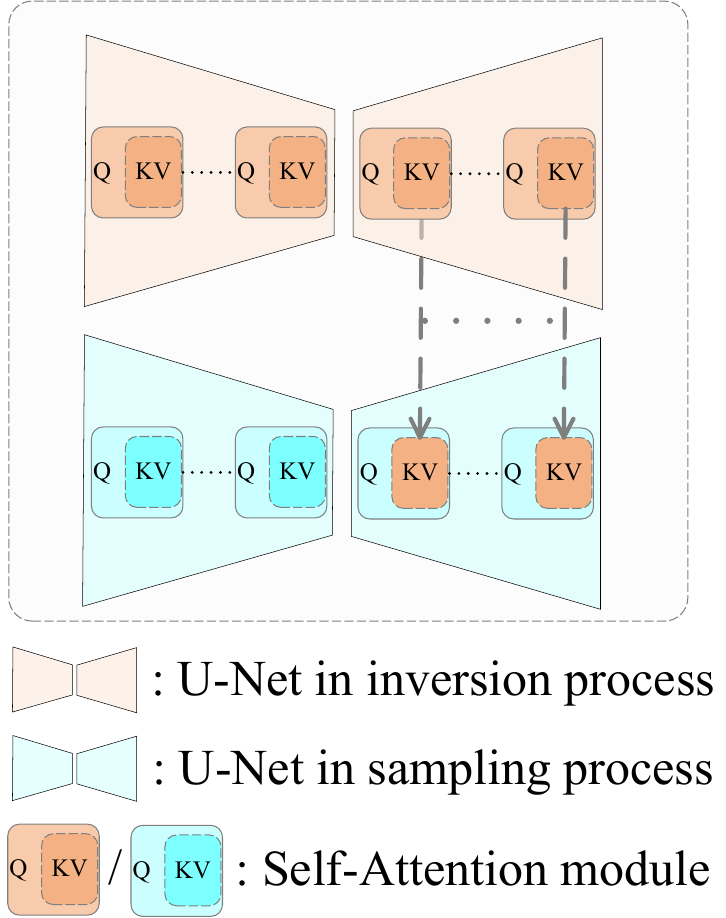}
        \caption{Illustration of consistency-preserving strategy.}
        \label{Consistency-Preserving}
    \end{minipage}
    \vspace{-0.5cm}
\end{figure}
To enhance the semantic integrity, BNNI interpolates points in null region using similar features from their neighboring areas in horizontal and vertical directions, thus ensuring the semantic integrity and enhancing the quality of drag editing. 
Let $\mathcal{N}$ be a point with coordinate $(x_{\mathcal{N}}, y_{\mathcal{N}})$ in null regions, we identify the nearest points of  $\mathcal{N}$ containing value in four directions: up, right, down, and left, as illustrated in \Cref{interpolation}, which are used as reference points for interpolation. Then, the interpolated value for null point $\mathcal{N}$ can be calculated as:
\begin{align}
\label{eq:9}
{\boldsymbol{z}'_t}_{(x_{\mathcal{N}}, y_{\mathcal{N}})}= \sum_{loc=u,r,d,l} w_{loc} \cdot {ref}_{loc} 
\end{align}
where ${ref}_{loc}$ denotes the value of reference point, and $loc$ indicates the direction, with $u$, $r$, $d$ and $l$ representing  up, right, down and left, respectively. $w_{loc}$ is the interpolation weight for each reference point, which is calculated based on its distance to $\mathcal{N}$, as follows:
\begin{align}
\label{eq:8}
w_{loc}=\frac{1/len_{loc}}{\sum_{loc=u,r,d,l}1/len_{loc}}
\end{align}
where $len_{loc}$ represents the distance between the reference point and $\mathcal{N}$.
Such that we can obtain the optimized noisy latent $\boldsymbol{z}'_t$ with complete semantic information by using BNNI to exploit similar semantic information from surrounding areas,  further enhancing the quality of the drag editing.
\subsection{Consistency-Preserving Strategy}
\label{sec:3.4}
Following~\cite{mou2023dragondiffusion, Cao2023MasaCtrlTM, shi2023dragdiffusion}, we introduce a consistency-preserving strategy to maintain the consistency between the edited image and the original image by adopting the semantic information of the original image (\ie, key and value pairs) saved in self-attention module during diffusion inversion to guide the diffusion sampling, as illustrated in \Cref{Consistency-Preserving}. 
Specifically, during the diffusion sampling, the calculation of self-attention $\text{Attention}_{\text{Sa}}$ within the upsampling process of the U-Net is as follows: 
\begin{align}
\text{Attention}_{\text{Sa}}(\boldsymbol{Q}_\text{Sa}, \boldsymbol{K}_\text{In}, \boldsymbol{V}_\text{In}) &= \text{softmax}( \frac{\boldsymbol{Q}_\text{Sa} \cdot \boldsymbol{K}_\text{In}}{\sqrt{d}} ) \cdot \boldsymbol{V}_\text{In}
\end{align}
where query $\boldsymbol{Q}_\text{Sa}$ is still used from diffusion sampling but key $\boldsymbol{K}_\text{In}$ and value $\boldsymbol{V}_\text{In}$ are correspondingly from diffusion inversion. Thus, the consistency-preserving strategy maintains the overall content consistency between the desired image and original image, ensuring the effect of drag-based editing.
\section{Experiments}
\label{sec:4}
\subsection{Qualitative Evaluation}
\label{Qualitative}
\begin{figure}[t]
    \vspace{-0.5cm}
    \centering
    \includegraphics[width=.97\textwidth]{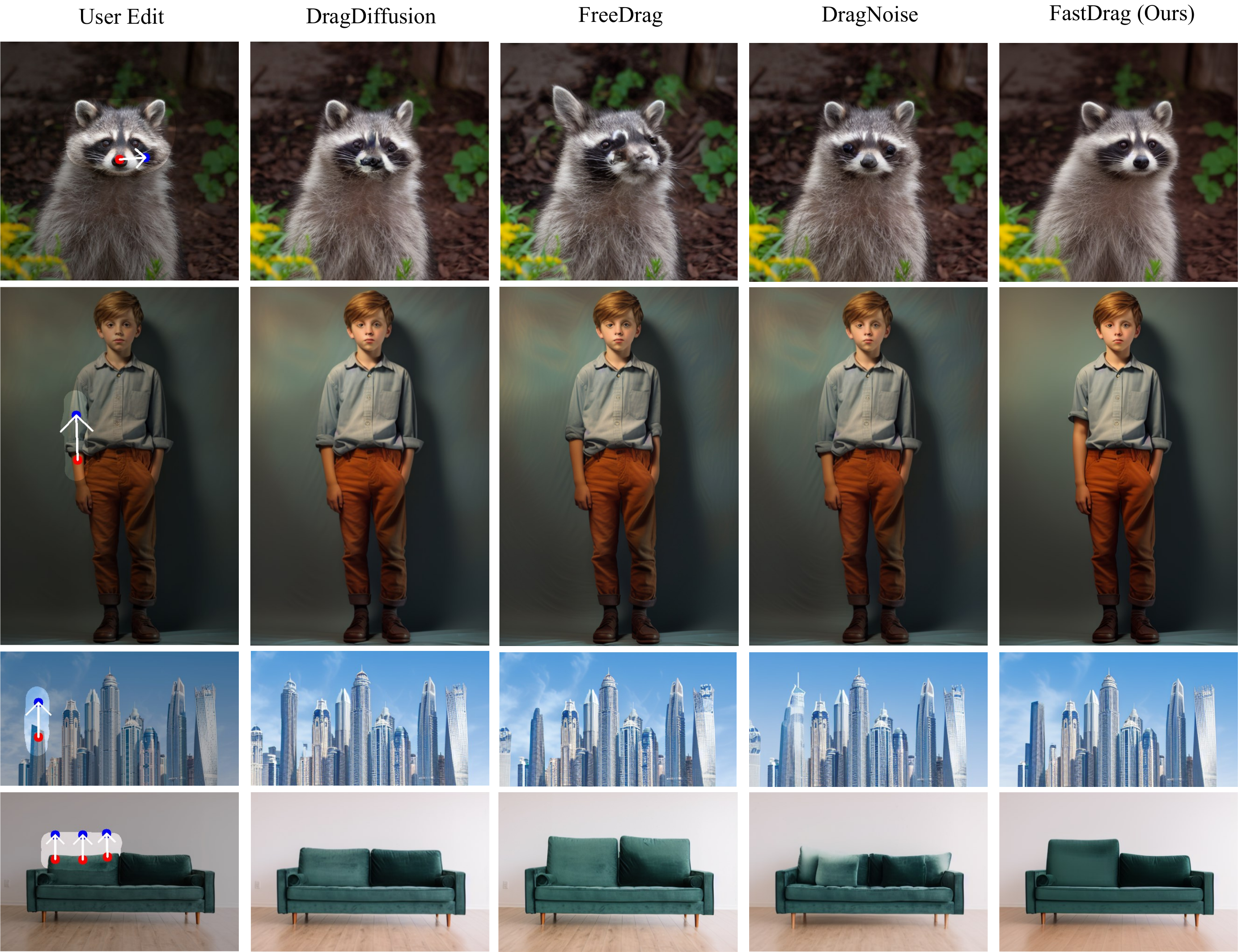}
    \caption{Illustration of qualitative comparison with the state-of-the-art methods.}
    \label{compare_results}
\end{figure}
We conduct experiments to demonstrate the drag effects of our FastDrag method, comparing it against state-of-the-art techniques such as DragDiffusion~\cite{shi2023dragdiffusion}, FreeDrag~\cite{ling2023freedrag}, and DragNoise~\cite{liu2024drag}. The qualitative comparison results are presented in \Cref{compare_results}. Notably, FastDrag maintains effective drag performance and high image quality even in images with complex textures, where $n$-step iterative methods typically falter. For instance, as shown in the first row of \Cref{compare_results}, FastDrag successfully rotates the face of an animal while preserving intricate fur textures and ensuring strong structural integrity. In contrast, methods like DragDiffusion and DragNoise fail to rotate the animal's face, and FreeDrag disrupts the facial structure.

In the stretching task, FastDrag outperforms all other methods, as shown in the second row of \Cref{compare_results}, where the goal is to move a sleeve to a higher position. The results show that other methods lack robustness to slight deviations in user dragging, where the drag point is slightly off the sleeve. Despite this, FastDrag accurately moves the sleeve to the desired height, understanding the underlying semantic intent of dragging the sleeve.

Additionally, we perform multi-point dragging experiments, illustrated in the fourth row of \Cref{compare_results}. Both DragDiffusion and DragNoise fail to stretch the back of the sofa, while FreeDrag incorrectly stretches unintended parts of the sofa. Through the LWF introduced in Sec.~\ref{sec:3.2.1}, FastDrag can manipulate all dragged points to their target locations while preserving the content in unmasked regions.
More results of FastDrag are illustrated in supplementary Sec.~\ref{sec:More Results}. 
\subsection{Quantitative Comparison}
\begin{figure}[t]
    \centering
    \begin{minipage}{.39\textwidth}
        \centering
        \includegraphics[width=1\textwidth]{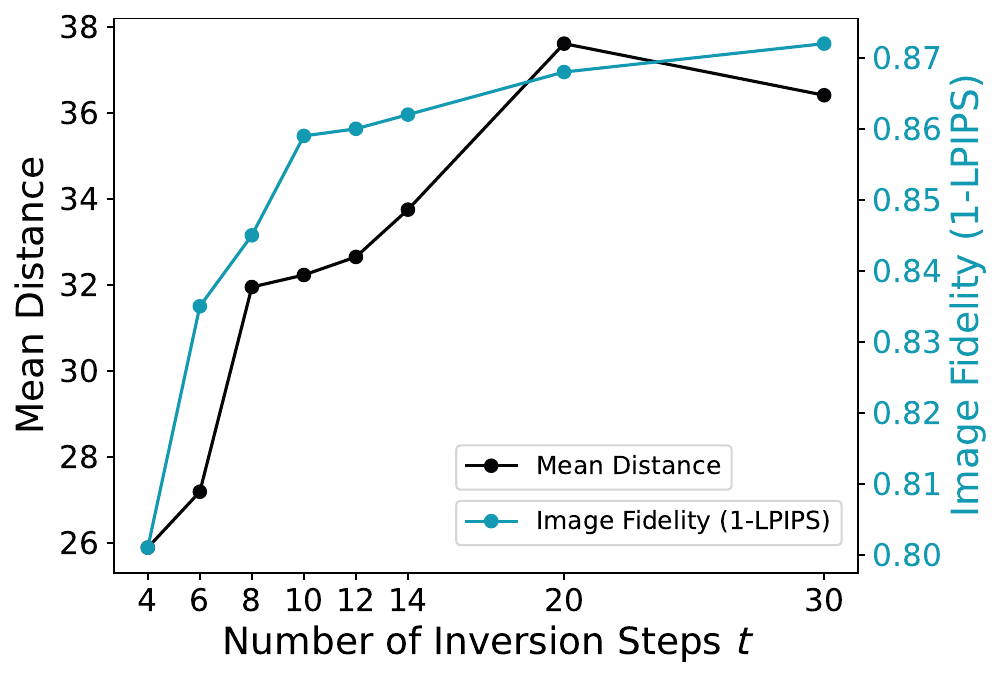}
        \caption{Ablation study on number of inversion steps in terms of quantitative metrics.}
        \label{ablution_inverse_chart}
    \end{minipage}
    \begin{minipage}{.58\textwidth}
        \captionsetup{type=table}
        \caption{Quantitative comparison with state-of-art methods on DragBench. Here, lower MD indicates more precise drag results, while higher 1-LPIPS reflects greater similarity between the generated and original images. The time metric represents the average time required per point based on RTX 3090. Preparation denotes LoRA training. ${\dagger}$ means FastDrag without LCM-equipped U-Net.}
        \resizebox{\linewidth}{!}{
        \begin{tabular}{cccccc}
        \toprule[1.5pt]
        \multicolumn{1}{c|}{ \multirow{2}*{Approach} }
        &\multicolumn{1}{c|}{ \multirow{2}*{$\mathrm{Venue}$} }
        &\multicolumn{1}{c}{ \multirow{2}*{$\mathrm{MD}\downarrow$} }
        &\multicolumn{1}{c}{ \multirow{2}*{$\mathrm{1-LPIPS}\uparrow$} }
        & \multicolumn{2}{c}{$\mathrm{Time}$}  \\
        \cline{5-6}
        \multicolumn{1}{c|}{}&\multicolumn{1}{c|}{}&\multicolumn{2}{c}{}&$\mathrm{Preparation}$ &$\mathrm{Editing(s)}$ \\
        \hline
        \hline
        DragDiffusion~\cite{shi2023dragdiffusion} &CVPR2024    & 33.70 & 0.89  &$1$~min~(LoRA)  &21.54\\
        DragNoise~\cite{liu2024drag} &CVPR2024     & 33.41 & 0.63 &$1$~min~(LoRA)  &20.41\\
        FreeDrag~\cite{ling2023freedrag} &CVPR2024         & 35.00 & 0.70  &$1$~min~(LoRA)  &52.63\\
        GoodDrag~\cite{zhang2024gooddrag} &arXiv2024         & 22.96 & 0.86 &$1$~min~(LoRA)   &45.83\\
        DiffEditor~\cite{mou2024diffeditor} &CVPR2024    & 28.46 & 0.89 &\XSolidBrush  &21.68\\
        FastDrag$^{\dagger}$(Ours) &  & 33.22 & 0.87  &\XSolidBrush  & 5.66\\
        FastDrag (Ours) &  & 32.23 & 0.86  &\XSolidBrush  & \textbf{3.12}\\
        \bottomrule[1.5pt]
        \end{tabular}
        \label{table:Quantitative}
        }
    \end{minipage}
\end{figure}

To better demonstrate the superiority of FastDrag, we conduct quantitative comparison using  DragBench dataset~\cite{shi2023dragdiffusion}, which consists of 205 different types of images with 349 pairs of handle and target points. Here, mean distance (MD)~\cite{pan2023drag} and image fidelity (IF)~\cite{Kawar_2023_CVPR} are employed as performance metrics, where MD evaluates the precision of drag editing, and IF measures the consistency between the generated and original images by averaging the learned perceptual image patch similarity (LPIPS)~\cite{Zhang_2018_CVPR}. Specifically, 1-LPIPS is employed as the IF metric in our experiment to facilitate comparison. In addition, we compare the average time required per point to demonstrate the time efficiency of our proposed FastDrag. The results are given in \Cref{table:Quantitative}.

Apart from~\cite{shi2023dragdiffusion},\cite{liu2024drag},\cite{ling2023freedrag}, two other state-of-the-art methods, \ie, GoodDrag~\cite{zhang2024gooddrag} and DiffEditor~\cite{mou2024diffeditor}, are also adopted for comparison, with DiffEditor being the current fastest drag-based editing method. Due to well-designed one-step warpage optimization and consistency-preserving strategy, our FastDrag does not require LoRA training preparation, resulting in significantly reduced time consumption (\ie, 3.12 seconds), which is nearly 700\% faster than DiffEditor (\ie, 21.68 seconds), and 2800\% faster than the typical baseline DragDiffusion (\ie, 1 min and 21.54 seconds). Moreover, even using standard U-Net without LCM, our method is still much faster than DiffEditor and far outperforms all other state-of-the-art methods. 
It is particularly noteworthy that, even with an A100 GPU, DiffEditor still requires 13.88 seconds according to \cite{mou2024diffeditor}, whereas FastDrag only requires 3.12 seconds on an RTX 3090.

In addition, our FastDrag also achieves competitive quantitative evaluation metrics  (\ie, IF and MD) comparable to the state-of-the-art methods, and even better drag editing quality, as illustrated in \Cref{compare_results}. These results demonstrate the effectiveness and superiority of our method.
\begin{figure}[!t]
    \centering
    \includegraphics[width=1\textwidth]{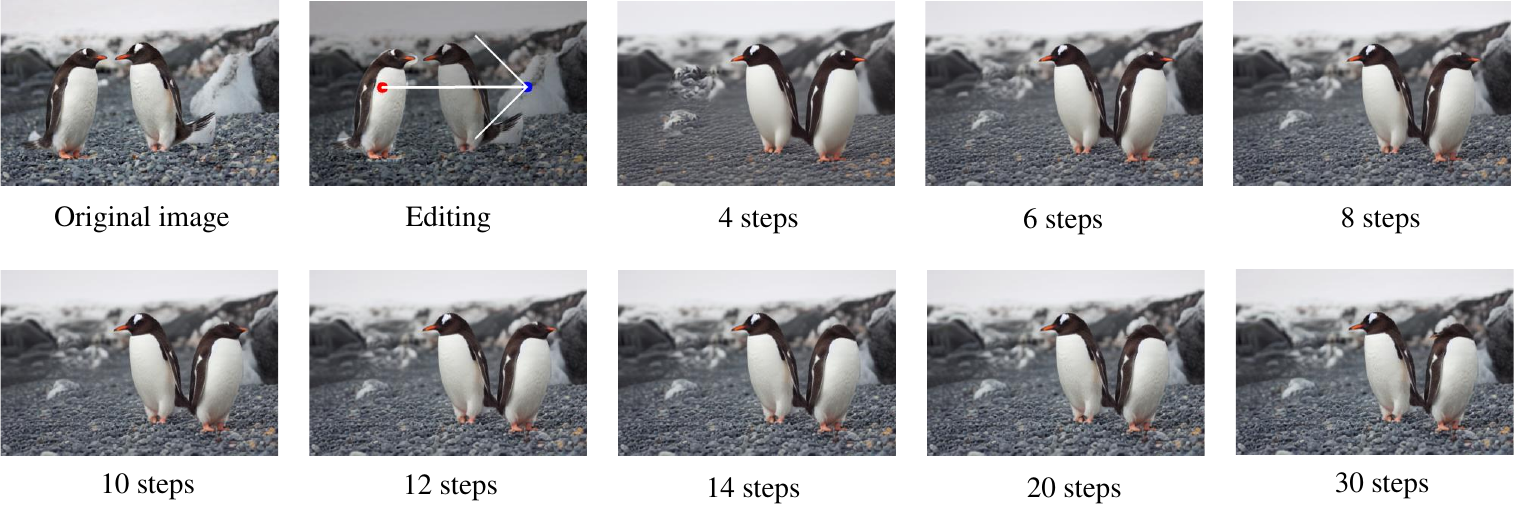}
    \caption{Ablation study on number of inversion steps in terms of drag effect.}
    \label{ablution_inversion}
\end{figure}
\subsection{Ablation Study}
\label{Ablation}
\noindent \textbf{Inversion Step:} To determine the number of inversion steps in diffusion inversion with LCM-equipped U-Net, 
we conduct an ablation experiment  with number of inversion steps set as $t$ = 4, 6, 8, 10, 12, 14, 20, and 30, where IF and MD are used to evaluate the balance between the consistency with original image and the  desired drag effects. The results are given in \Cref{ablution_inversion} and \Cref{ablution_inverse_chart}, where we can see that when $t < 6$, the generated images lack sufficient detail to accurately reconstruct the original images. Conversely, when $t > 6$, it can successfully recover complex details such as intricate fur textures and dense stone while maintaining high image quality. 
However, when $t > 14$, some image details lost, which negatively impacts the effectiveness of the drag effect. 
By comprehensive evaluation of both the drag effect and the similarity to the original images, we select 10 as the number of inversion steps for our method with LCM to balance the drag effect. 

\begin{figure}[!t]
    \centering
    \includegraphics[width=1\textwidth]{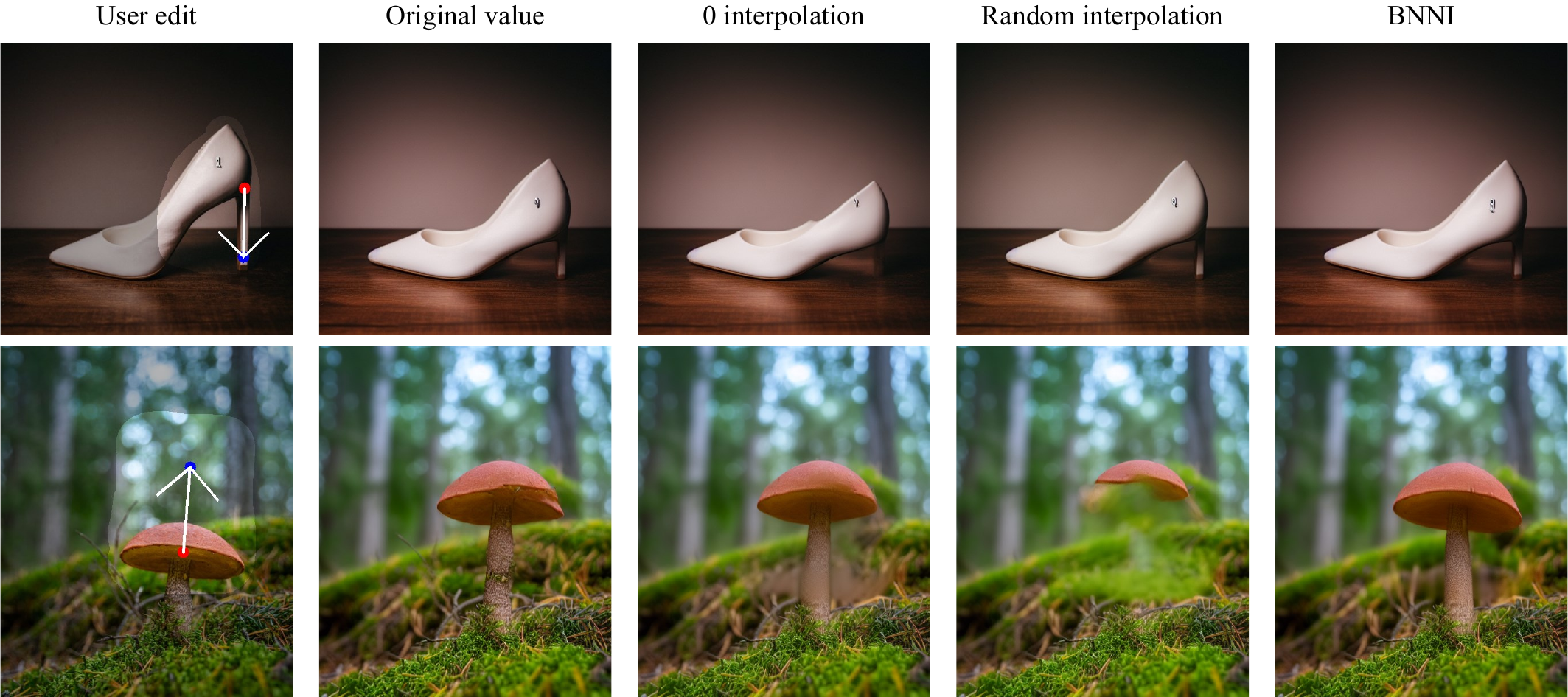}
    \caption{Ablation study on  bilateral nearest neighbor interpolation.}
    \label{ablution_BNNI}
   \vspace{-0.3cm}
\end{figure}

\noindent \textbf{BNNI:} To demonstrate the effectiveness of BNNI, we compare it with several interpolation methods on null point $\mathcal{N}$, including maintaining the original value of this position, interpolation by zero-value, and interpolation by random noise, denoted as ``original value'', ``0 interpolation'', and ``random interpolation'', respectively. The results are given in \Cref{ablution_BNNI}, where we can see that, 
by effectively utilizing surrounding feature values to interpolate null points, BNNI can address semantic losses, and enhance the quality of the drag editing.

\begin{figure}[ht]
    \centering
    \includegraphics[width=1\textwidth]{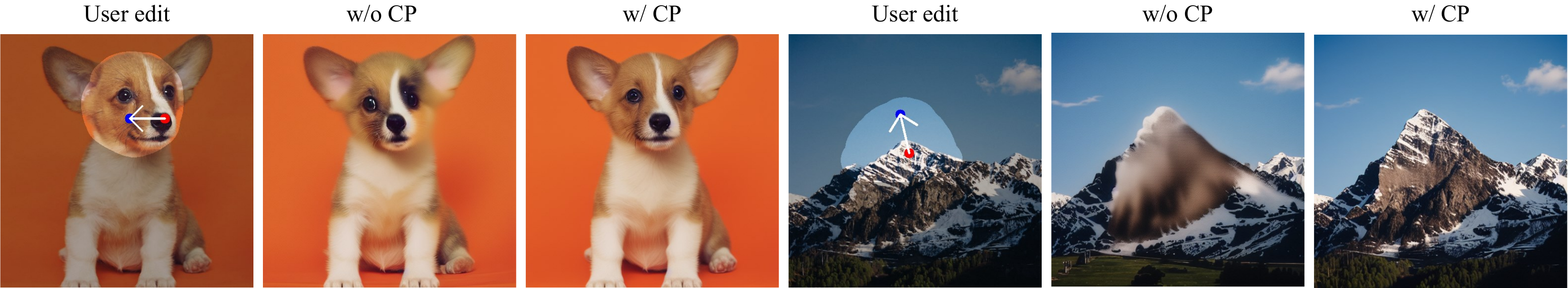}
    \caption{Ablation study on consistency-preserving strategy.}
    \label{fig:anlution_kv}
\end{figure}

\noindent \textbf{Consistency-Preserving:}
We also conduct an experiment to validate the effectiveness of the consistency-preserving strategy in maintaining image consistency. The results are illustrated in \Cref{fig:anlution_kv}, where ``w/ CP'' and ``w/o CP''  denote our FastDrag with and without using consistency-preserving strategy, respectively. It is obviously that our method with consistency-preserving strategy can effectively preserve image consistency, resulting in better drag editing effect.
%
%

\section{Limitations}
\label{sec:Limitations}
\begin{figure}[!t]
\centering
\vspace{-0.2cm}
\includegraphics[width=0.94\linewidth]{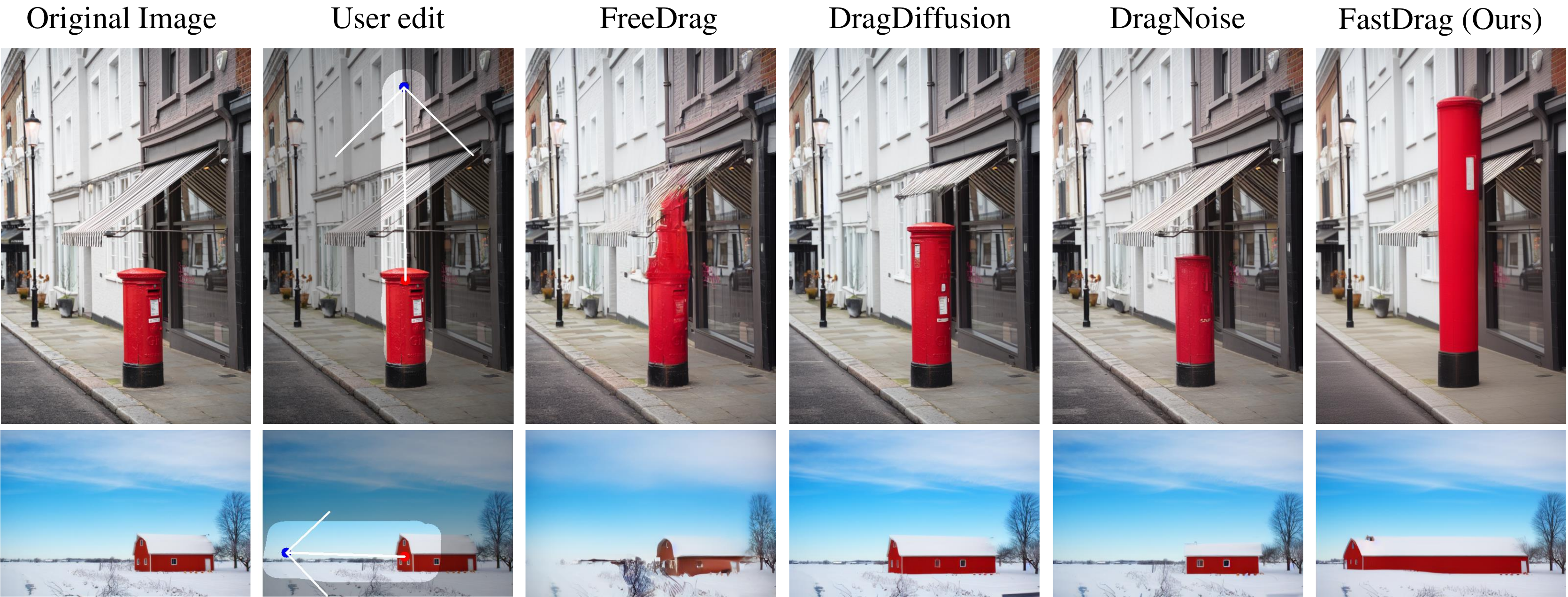}
\caption{\label{long}
Illustration of failure cases for limitation analysis under extremely long-distance drag editing. 
Our FastDrag method may lose some detailed information in these cases but still achieves better editing performance compared to state-of-the-art (SOTA) methods.
}
\end{figure}

\begin{figure}[!t]
\centering
\includegraphics[width=0.95\linewidth]{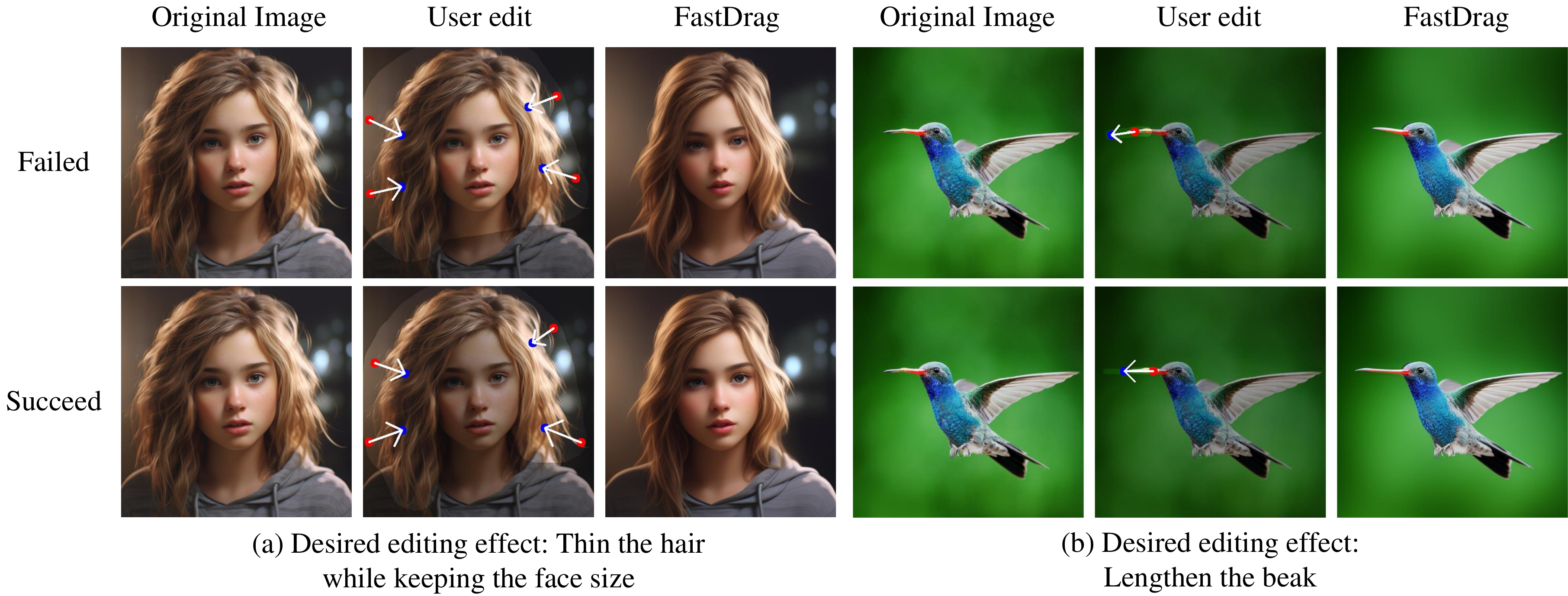}
\caption{\label{fail}Illustration of the limitation analysis with failed and successful drag editing for highly relying on precise drag instruction. (a) It is best to exclude the face from the mask region. (b) The handle point should ideally be placed where the ``beak" feature is more prominent.}
\vspace{-0.2cm}
\end{figure}

Despite FastDrag’s impressive editing speed compared to SOTA methods, it shares some common limitations. 
1)~\textbf{Overly Smooth and Finer Details Loss:} Similar to other diffusion-based methods~\cite{cho2024noise, shi2023dragdiffusion}, FastDrag occasionally loses fine textures from the original images, as shown in~\Cref{compare_results}, row 4. Despite this, FastDrag outperforms other methods in speed and overall performance. 
2)~\textbf{Extremely Long-distance Drag Editing:} In such case, object details may be lost due to the lower-dimensional latent space, in which significant changes in detail~(i.e., long-drag editing) can disrupt the semantics, making it harder to preserve all details. Nevertheless, FastDrag handles long-distance editing better than other SOTA methods, as illustrated in~\Cref{long}, where our method successfully achieves long-distance drag editing that others fail to achieve.
3)~\textbf{Highly Relying on Precise Drag Instruction:} Achieving optimal results depends heavily on clear drag instructions. As with other SOTA methods, precise input, such as excluding irrelevant areas from the mask (e.g., the face in~\Cref{fail}, row 2) or correctly placing the handle point (e.g., beak in~\Cref{fail}), is essential for better performance.

\section{Conclusion}
This paper has presented a novel drag-based image editing method, \ie, FastDrag, which achieved faster image editing speeds than other existing methods. By proposing one-step warpage optimization and BNNI strategy, our approach achieves high-quality image editing according to the drag instructions in a very short period of time. Additionally, through the consistency-preserving strategy, it ensures the consistency of the generated image with the original image. Moving forward, we plan to continue refining and expanding our approach to further enhance its capabilities and applications.

\section{Acknowledgments}
This work was partly supported by Beijing Nova Program (20230484261).

{\small
\bibliographystyle{ieee_fullname}
\bibliography{egbib}
}
\appendix
\newpage
\section{Supplementary Experiments}
\label{sec:Supplementary experiment}
We conduct supplementary quantitative experiments on BNNI and consistency-preserving strategies to further validate their effectiveness. The quantitative metrics used are consistent with those described in Sec.~\ref{Qualitative}.

\begin{figure}[ht]
    \centering
    \begin{minipage}{.48\textwidth}
        \centering
        \includegraphics[width=1\textwidth]{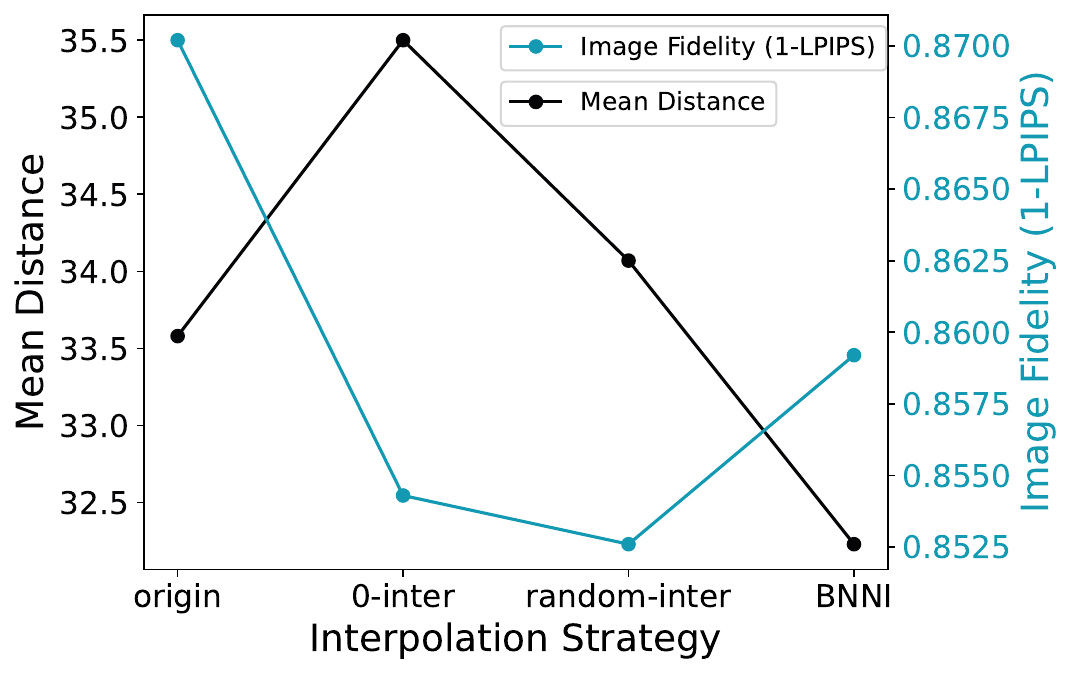}
        \caption{Ablation study on BNNI in terms of quantitative metrics.}
        \label{ablution_BNNI_chart}
    \end{minipage}
    \hspace{0.2cm}
    \begin{minipage}{.45\textwidth}
        \centering
        \includegraphics[width=1\textwidth]{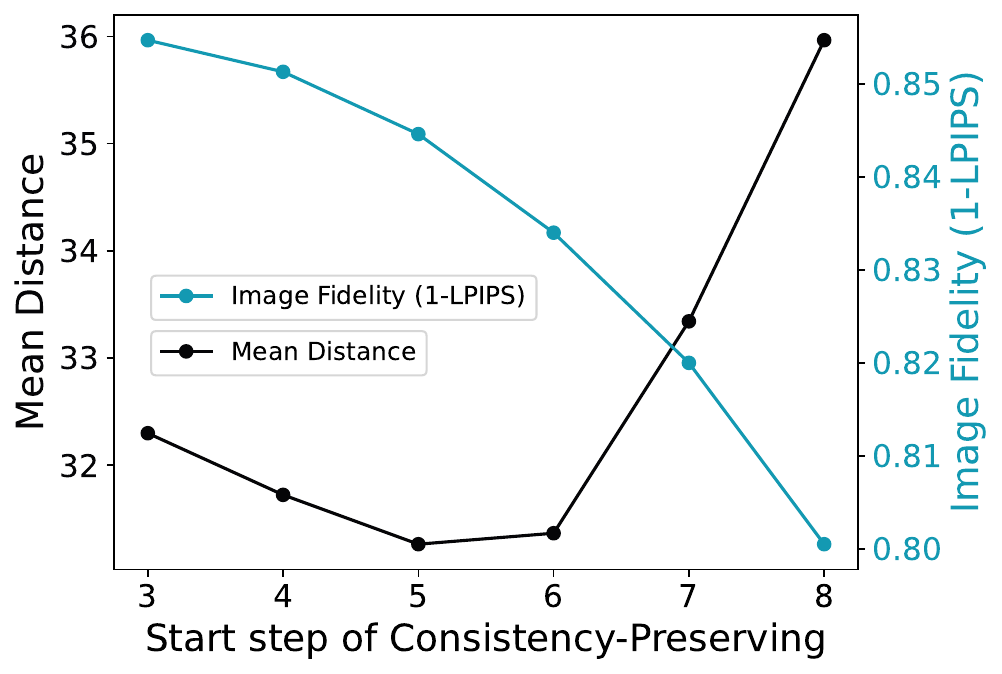}
        \caption{Ablation study on consistency-preserving in terms of quantitative metrics.}
        \label{ablution_kv_step_chart}
    \end{minipage}
    \vspace{-.25cm}
\end{figure}

\noindent \textbf{BNNI:} Following the setup in Sec.\ref{Ablation}, we compare it with several interpolation methods on null point $\mathcal{N}$, including maintaining the original value of this position, interpolation by zero-value, and interpolation by random noise, denoted as ``origin'', ``0-inter'', and ``random-inter'', respectively. As illustrated in~\Cref{ablution_BNNI_chart}, FastDrag with BNNI achieves the best MD levels compared to other interpolation methods, while its IF is second only to ``origin''. However, ``origin'' can lead to negative drag effects, as shown in~\Cref{ablution_BNNI}. Therefore, by effectively utilizing surrounding feature values to interpolate null points, BNNI can address semantic losses and enhance the quality of drag editing.

\noindent \textbf{Consistency-Preserving:} 
We also conduct experiments to assess the impact of initiating the consistency-preserving strategy at different sampling steps. The results, as shown in~\Cref{ablution_kv_step_chart}, indicate that as the starting step increases ( \ie, the frequency of key and value replacements decreases), the IF decreases, leading to poorer image consistency. Meanwhile, the MD initially decreases and then increases as the starting step increases. It is evident that consistency-preserving strategy can effectively maintain the consistency between the generated images and the original images. 

\begin{wrapfigure}{r}{0.5\textwidth}
\setlength{\abovecaptionskip}{0.cm}
\setlength{\belowcaptionskip}{0.cm}
    \vspace{-1cm}
    \begin{center}
    \includegraphics[width=.45\textwidth]{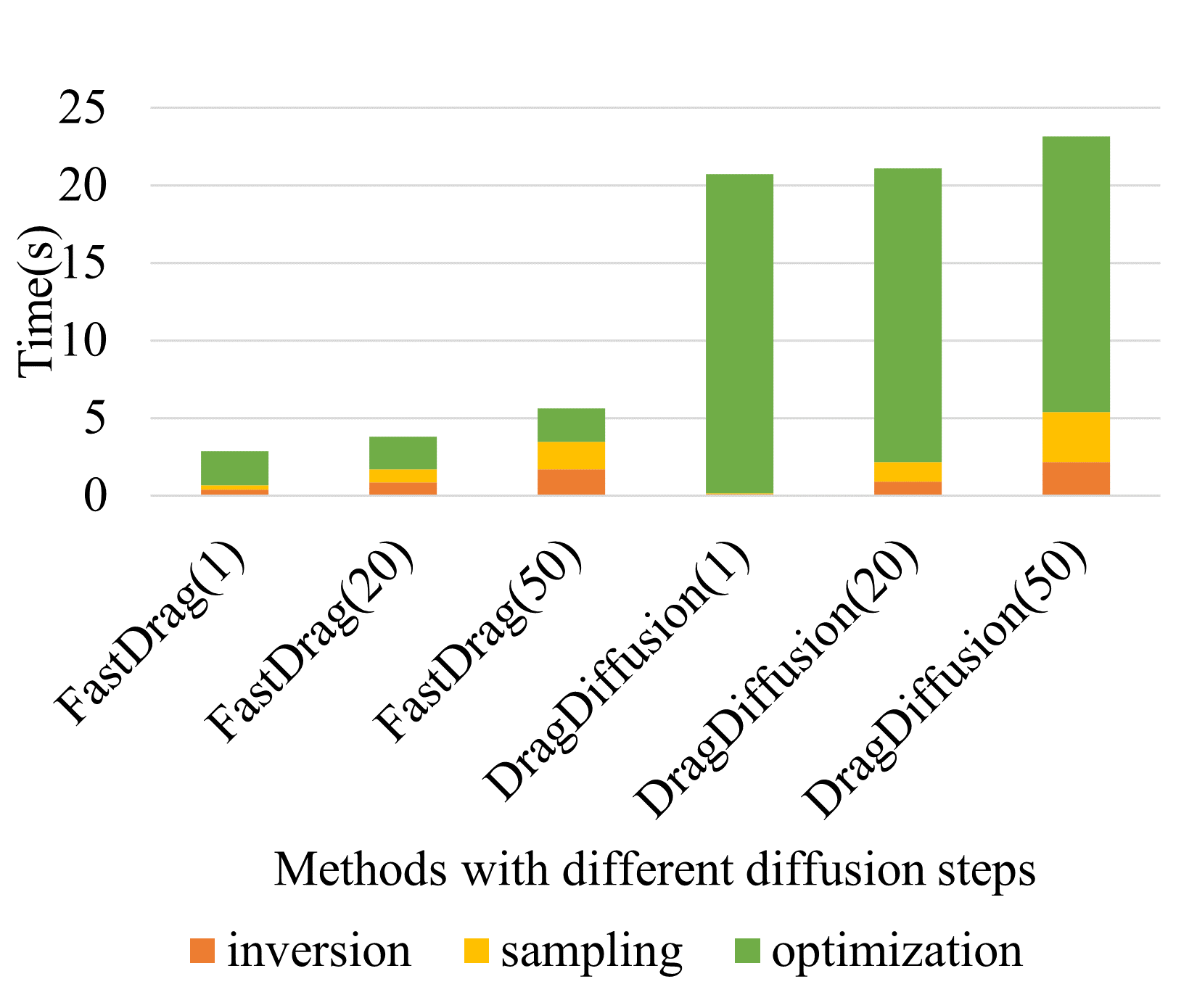}
    \end{center}
    \caption{Overall editing time comparison with different diffusion steps between FastDrag and DragDiffusion. All experiments are conducted on RTX 3090 with diffusion step set as 1, 20, and 50 respectively. Optimization means latent optimization. }
    \label{single diffusion}
    \vspace{-1cm}
\end{wrapfigure}

\section{Single-step Diffusion}
When integrating DragDiffusion with a single-step diffusion model, the editing time is still much longer than that of FastDrag. For DragDiffusion and FastDrag under diffusion steps of 1, 20, and 50, we calculate the time required for inversion, sampling, and latent optimization respectively. The results provided in~\Cref{single diffusion} show that even with a single diffusion step (i.e., diffusion step set as 1), DragDiffusion still requires significantly more time (20.7 seconds) compared to FastDrag (2.88 seconds). 

In addition, as observed in~\Cref{single diffusion}, DragDiffusion spends significantly more time on latent optimization compared to diffusion inversion and sampling. Therefore, reducing the time spent on latent optimization is crucial for minimizing overall editing time, which is precisely what FastDrag accomplishes.


\section{Statistical Rigor}
\label{Statistical Rigor}
To further validate the superiority of our work and to achieve statistical rigor, we conducted an additional experiment by repeating our experiment 10 times under the same experimental settings. We observed that the variances of the performance metrics obtained from 10 realizations of our FastDrag are MD (0.000404), 1-LPIPS (9.44E-11), and Time (0.018), all of which fall within a reasonable range. These statistical results further demonstrate the effectiveness and stability of our method for drag editing.

\section{Implementation Details}
\label{sec:4.1}
We utilize a pretrained LDM (\ie, Stable Diffusion 1.5 \cite{Rombach_2022_CVPR}) as our diffusion model, where the U-Net structure is adapted with LCM-distilled weights from Stable Diffusion 1.5. It is worth emphasizing that the U-Net structure used in our model is widely used in image generation methods~\cite{shi2023dragdiffusion, ling2023freedrag, liu2024drag, zhang2024gooddrag, Cui2024StableDragSD}. Unless otherwise specified, the default setting for inversion and sampling step is 10.   
Following DragDiffusion~\cite{shi2023dragdiffusion}, classifier-free guidance (CFG)~\cite{ho2021classifierfree} is not applied in diffusion model, and we optimize the diffusion latent at the 7th step. All other configurations follow that used in DragDiffusion. Our experiments are conducted on an RTX 3090 GPU with 24G memory.

For the special application of drag-based editing, \ie, object moving, as shown in~\Cref{ablution_inversion}, significant null region may be left at the original position of the object due to long-distance relocation, posing challenges for BNNI. To ensure semantic integrity in the image and facilitate user interaction, we adopted two straightforward strategies. For first strategy, specifically, we introduce a parameter $r$, set as 2, centered at the unique target point $e_1$, defining a rectangular area with dimensions $2r$. We then extract the noised-latent representation within this rectangular area and fill it into the mask, effectively restoring the semantics at the original position of the object. For second strategy, we just maintain the original semantic of original position to avoid null region. Under second strategy, object moving will produce the effect of object replication, as shown in the third row of~\Cref{More Results}.
\section{More Results}
\label{sec:More Results}
\begin{figure}[!htp]
    \centering
    \includegraphics[width=.97\textwidth]{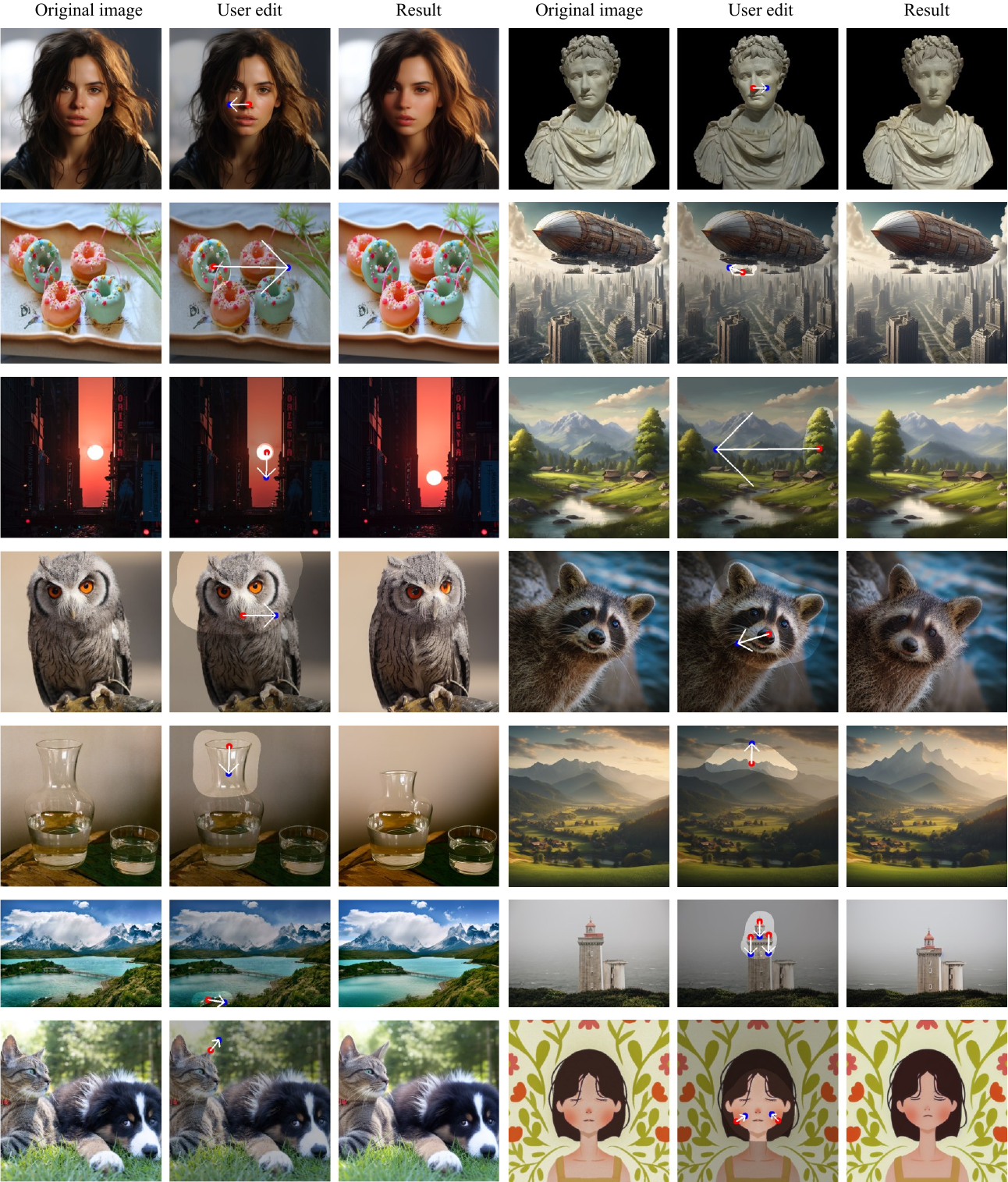}
    \caption{More visualized results of FastDrag.}
    \label{More Results}
\end{figure}

We apply FastDrag to drag-based image editing in various scenarios, including face rotation, object movement, object stretching, object shrinking and so on. The experimental results, as shown in~\Cref{More Results}, demonstrate that FastDrag achieves excellent drag-based effects across multiple scenarios.

\section{Societal Impacts}
\label{Societal Impacts}

FastDrag has the potential to bring about several positive societal impacts. Firstly, they offer intuitive and efficient image editing tools, catering to artists, designers, and creators, thereby fostering creativity and innovation. Secondly, their user-friendly nature simplifies the image editing process, increasing accessibility and participation among a wider audience. In addition, our FastDrag improves efficiency and saves the user's time and effort, thus increasing productivity. Moreover, the innovative and flexible nature of these methods opens up possibilities for various applications, spanning art creation, design, education, and training.

However, along with these positive aspects, FastDrag can also have certain negative societal implications. They may be exploited by unethical individuals or organizations to propagate misinformation and fake imagery, potentially contributing to the spread of false news and undermining societal trust. Furthermore, widespread use of image editing tools may encroach upon individual privacy rights, particularly when unauthorized information or imagery is manipulated. Moreover, inappropriate or irresponsible image editing practices could lead to social injustices and imbalances, such as distorting facts or misleading the public, thereby influencing public opinion and policies negatively.


\end{document}